\def\year{2022}\relax
\documentclass[letterpaper]{article} 
\usepackage{aaai22}  
\usepackage{times}  
\usepackage{helvet}  
\usepackage{booktabs}
\usepackage{courier}  
\usepackage[hyphens]{url}  
\usepackage{multirow}
\usepackage{multicol}
\usepackage{amsmath}
\usepackage{graphicx} 
\usepackage{natbib}  
\usepackage{caption} 
\DeclareCaptionStyle{ruled}{labelfont=normalfont,labelsep=colon,strut=off} 
\usepackage{threeparttable}
\usepackage{algorithm}
\usepackage{algorithmic}

\usepackage{newfloat}
\usepackage{listings}
\floatstyle{ruled}
\newfloat{listing}{tb}{lst}{}
\floatname{listing}{Listing}

\usepackage{xcolor}
\usepackage{xspace}
\usepackage{bm}
\usepackage{amsmath}

\usepackage{xcolor}
\usepackage{color, soul}

\title{From Good to Best: Two-Stage Training for Cross-lingual Machine Reading Comprehension}
\author{
   Nuo Chen$^{1}$\footnote{Work done during internship at Microsoft STCA.},
   Linjun Shou$^2$,
   Min Gong$^2$,
   Jian Pei$^3$,
   Daxin Jiang$^2$\footnote{Corresponding Author}
   }
  \affiliations{
$^1$ADSPLAB, School of ECE, Peking University, Shenzhen, China\\
$^2$NLP Group, Microsoft STCA\\
$^3$School of Computing Science, Simon Fraser University \\
nuochen@pku.edu.cn,
\{lisho,migon\}@microsoft.com,
jpei@cs.sfu.ca
}
\usepackage{bibentry}

\begin{document}

\maketitle

\begin{abstract}


Cross-lingual Machine Reading Comprehension (xMRC) is challenging due to the lack of training data in low-resource languages.  The recent approaches use training data only in a resource-rich language like English to fine-tune large-scale cross-lingual pre-trained language models.
Due to the big difference between languages, a model fine-tuned only by a source language may not perform well for target languages. Interestingly, we observe that while the top-1 results predicted by the previous approaches may often fail to hit the ground-truth answers, the correct answers are often contained in the top-k predicted results. 
Based on this observation, we develop a two-stage approach to enhance the model performance. The first stage targets at recall: we design a hard-learning (HL) algorithm to maximize the likelihood that the top-k predictions contain the accurate answer. The second stage focuses on precision:  an answer-aware contrastive learning (AA-CL) mechanism is developed to learn the fine difference between the accurate answer and other candidates. Our extensive experiments show that our model significantly outperforms a series of strong baselines on two cross-lingual MRC benchmark datasets.


\end{abstract}
\section{Introduction}


Machine Reading Comprehension (MRC) has been intensively studied in the Natural Language Understanding community in the past years~\cite{DBLP:conf/emnlp/RajpurkarZLL16,DBLP:conf/iclr/YuDLZ00L18, DBLP:conf/icassp/ChenLYZZ21,DBLP:conf/ijcai/YouCZ21,DBLP:conf/iclr/SeoKFH17,DBLP:conf/wsdm/LiangSPGZJ21,DBLP:conf/emnlp/RajpurkarZLL16, DBLP:conf/acl/RajpurkarJL18,DBLP:journals/tacl/ReddyCM19,DBLP:journals/corr/abs-2010-08923,DBLP:journals/corr/abs-2010-11066}. When scaling out MRC to multiple languages, i.e., the task of cross-lingual MRC or xMRC for short, one challenge is the lack of training data in low-resource languages, where no training examples are available. To tackle this challenge, recent approaches build on large-scale cross-lingual pre-trained language models, such as mBERT~\cite{pires2019multilingual} and XLM-R~\cite{DBLP:journals/corr/abs-1911-02116}. These pre-trained models map the representations of different languages into a universal semantic space, where the expressions in different languages are represented close to each if they have similar meanings. The cross-lingual pre-trained models are then fine-tuned by training data only in a source language (e.g., English), and finally applied to various target languages. This approach has shown promising results on tasks such as entity recognition~\cite{DBLP:conf/wsdm/LiangSPGZJ21,DBLP:conf/acl/KruengkraiNMB20}, question answering~\cite{DBLP:journals/corr/abs-1809-03275,DBLP:conf/naacl/ZhouGSZJ21}, as well as xMRC~\cite{DBLP:conf/acl/YuanSBGLDFJ20,DBLP:conf/wsdm/LiangSPGZJ21}. However, due to the big difference between languages, the model fine-tuned only on the source language may not perform well on target languages.


\begin{table}[]
\centering
\small
\begin{tabular}{ccccccc}

\toprule

\textbf{Language}  & en & es  & de & ar & hi & vi \\ \midrule
\textbf{EM score} & 64.24 & 48.30 & 46.43 & 35.14 & 41.93 & 42.36 \\ \bottomrule
\end{tabular}
\caption{EM scores across different languages on MLQA dataset.}
\vspace{-10pt}
\label{table11}

\end{table}


Table~\ref{table11} is the results from our empirical study by applying the previous approach. To be more specific, we use English data to fine-tune the cross-lingual XLM-R model~\cite{DBLP:journals/corr/abs-1911-02116} on MLQA dataset~\cite{DBLP:conf/acl/LewisORRS20}. The numbers in the table are exact match (EM) scores, which is a widely-adopted metric for MRC task to evaluate the match degree between the model predicted result and the ground-truth answer. For each case, we take the top-1 output from the model as the predicted result. From Table~\ref{table11}, we clearly see the result on English is much better than that on other languages. The reason is that the model is fine-tuned by only English training data. At the same time, the model can still achieve 35 to 48 EM scores on non-English languages even though it has never been trained by any examples from those languages. This suggests that the model has inherited certain extent of language transfer capability from the cross-lingual pre-trained models.

We then extend the set of model predicted results by including top-k outputs from the model. That is, we regard the model ``successfully'' predicts the answer if any one in the top-k outputs matches the ground-truth answer. The modified EM scores with varying numbers of k are illustrated in Table~\ref{table12} (note the numbers in the column of ``Top-1''  are just those in Table~\ref{table11}). From Table~\ref{table12}, we can see the scores have substantial gains for all languages when we increase k. The gain in English is the smallest (around 10 points when k=10 vs. k=1), since the model has been well fine-tuned on this language by the native training examples. In other languages, the gains are much larger, i.e., more than 20 points when k=10 vs. k=1. This observation discloses huge potential in the top-k results. Intuitively, it suggests that the model has been empowered with the ability to roughly distinguish good results with bad ones. However, without sufficient training examples, it is not powerful enough to rank the most accurate result at the top-1 position.

The analysis of results in Table~\ref{table12} motivates us to decompose the training for xMRC model into two stages. The first stage targets at recall at top-k, which maximizes the likelihood for the accurate answer to be included in the set of top-k results. For this purpose, we design a hard-learning (HL) algorithm to learn the margin between good answers and bad ones. The second stage focuses on precision at top-1. We propose an answer-aware contrastive learning (AA-CL) mechanism to enable the model to further distinguish the accurate answer from the other candidates. Instead of selecting random or in-batch negatives, AA-CL constructs hard-negatives using the candidate that is most similar with (but not equal to) the ground-truth answer in the top-k prediction set at each update. Such hard-negatives help the model improve precision at top-1. 

\begin{table}[]
\centering

\begin{tabular}{ccccc}

\toprule
\textbf{Language}  & Top-1 & Top-3  & Top-5 & Top-10 \\ \midrule

en & 64.24 & 73.06 & 75.69 & 75.76 \\
es & 48.30 & 60.32 & 66.04  & 71.18 \\
de & 46.43 & 60.99  & 67.13 & 72.17 \\
ar & 35.14 & 48.24 & 52.33 & 57.13 \\
 hi& 41.93 & 57.70 & 63.32 & 70.15 \\ 
 vi& 42.36 & 58.25 & 61.98 & 66.06 \\  \bottomrule
\end{tabular}
\caption{EM scores among different top-k answer predictions on MLQA dataset, respectively.}
\vspace{-10pt}
\label{table12}

\end{table}


Our technical contributions are summarized as follows:
\begin{itemize}

    \item We conduct an in-depth study on the xMRC task, make interesting observations, and design a novel two-stage approach based on the observations. 
    
    \item We carry out extensive experiments and verify that our approach significantly surpasses previous state-of-the-art cross-lingual PLMs on two popular benchmarks.
\end{itemize}

The rest of the paper is organized as follows: We first review related work in Section 2, and then describe our proposed method detaily in Section 3. We report the extensive experiment results in Section 4, and further conduct ablation studies and analysis in Section 5. Finally, we conclude the paper in Section 6. 

\section{Related Work}
\paragraph{Cross-Lingual Machine Reading Comprehension} 
Recently, a considerable amount of literature has been published on cross-lingual machine reading comprehension (xMRC). A naive but efficient way is based on the machine translation system, which translates the training data in a rich-resource languages into other low-resource target language. Given the translated data, \citet{DBLP:conf/emnlp/CuiCLQWH19} proposed to use back-translation for xMRC. \citet{DBLP:journals/corr/abs-1905-11471} developed a new translation-based data augmentation method for multilingual  training. Unfortunately, all these methods heavily rely on the high-quality translation systems. On the other line, a school of approaches \cite{DBLP:conf/emnlp/HuangLDGSJZ19,DBLP:conf/emnlp/LiangDGWGQGSJCF20,DBLP:journals/corr/abs-1911-02116} based on large-scale multilingual pre-trained language models (PLMs) have been proposed. And a series of experiments prove that these PLMs can achieve superior performances even if in zero-shot or few-shot setting. More recently, several efforts have been made to further improve the PLMs performance in xMRC. To address the answer boundary problem in low-resource languages,  \citet{DBLP:conf/acl/YuanSBGLDFJ20} proposed several auxiliary tasks on top of PLMs so as to improve the model performance. Following the line, \citet{DBLP:conf/wsdm/LiangSPGZJ21} presented a calibration neural network in a pre-training manner. Nevertheless, none of these studies
explore to utilize top-k  predictions from a base model as weak supervisions to train more robust models for xMRC.

\paragraph{Contrastive Learning} Nowadays, Contrastive learning \cite{hadsell2006dimensionality} has been seen as a promising way to build on learning effective representations by pulling together semantically close neighbors (\textit{positive}) in a shared embedding space, and pushing apart non-neighbors (\textit{negatives}). Contrastive learning objective has been particularly successful in different contexts of vision, language, and speech \cite{kharitonov2021data,he2020momentum,DBLP:journals/corr/abs-2104-08821,DBLP:conf/emnlp/YouCZ21,you2021momentum,you2021simcvd}. \citet{DBLP:journals/corr/abs-2012-15466} proposed several sentence-level augmentation strategies to obtain a noise-invariant representation for down-stream tasks, such as text similarity and sentiment classification. Most recently,  \citet{DBLP:journals/corr/abs-2104-08821} developed a simple contrasive learning method via using  dropout \cite{DBLP:journals/jmlr/SrivastavaHKSS14} as noise. Concretely, they passed the same sentence into the PLMs twice and obtained positive pairs by applying dropout masks randomly. Although contrastive learning achieves significantly success in various natural language processing tasks, the context of question answering is less explored by research communities, especially for MRC. In this paper, we focus on a more challenge scenario: we propose AA-CL to leverage  hard-negatives from highly confident predictions for xMRC, in which the hard-negatives are consistently updated during training.

\begin{figure*}
    \centering
    \includegraphics[width=0.9\linewidth]{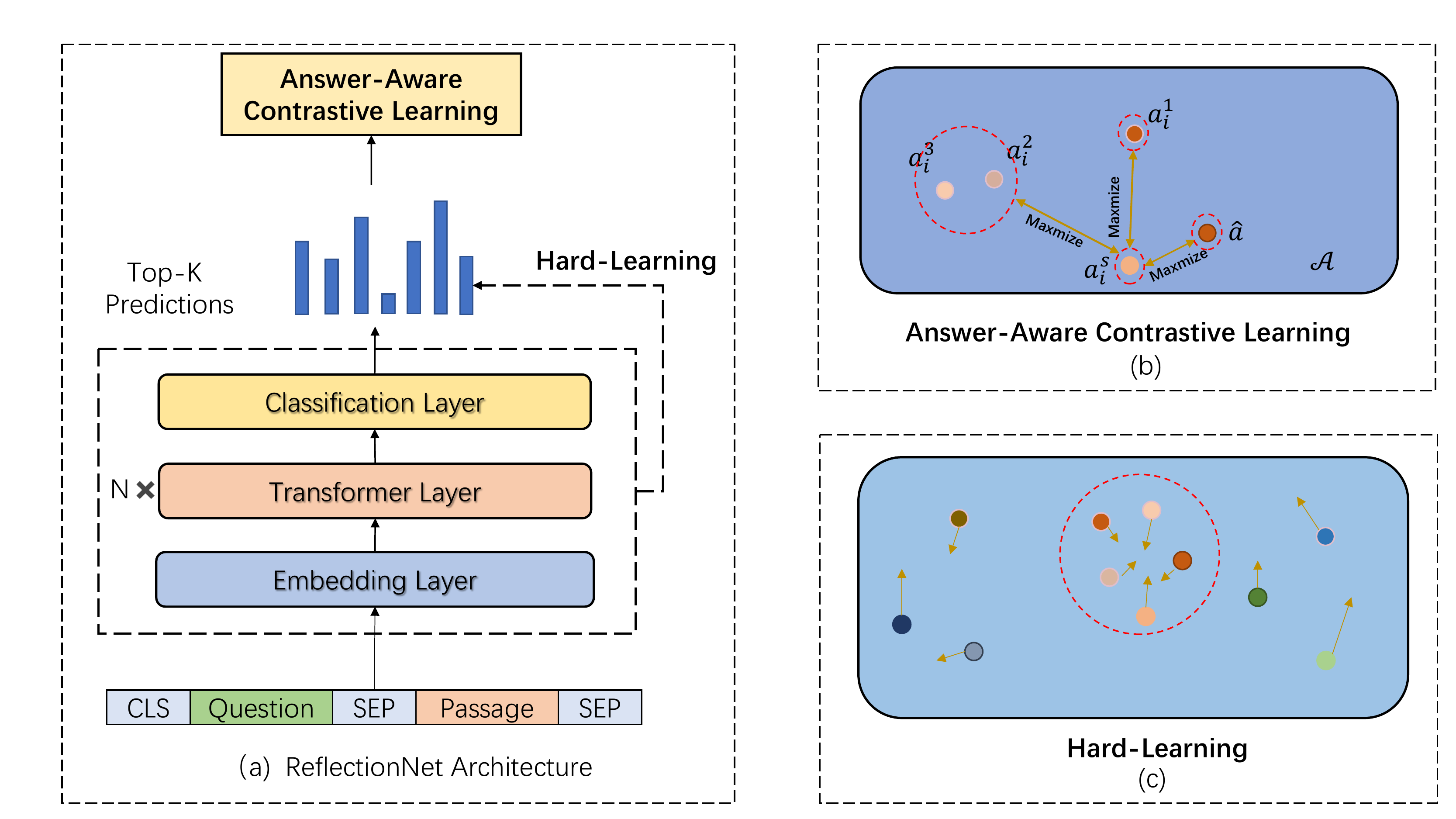}
    \caption{Overview of our proposed method.}
    \label{modeloverview}
\end{figure*}

\section{Model}

In this section, we aim to describe our proposed methods (See Figure \ref{modeloverview}) in detail. First, we introduce the problem formulation of xMRC. Then we describe the baseline model of our work. Last, we illustrate  the hard-learning (HL) algorithm and answer-aware contrastive learning (AA-CL) sequentially.

\subsection{Problem Formulation}
The problem of xMRC studied in this paper can be formulated as follows. In this work, assume that our labeled data collection $\mathcal{D}{_s} \in\{q_i,p_i,a_i\}_{i}^{N}$ in a source language (rich-resource). Specifically,  $\{q_i,p_i,a_i\}$ denotes the $i$-th triplets of $\{question, passage, answer\}$ in the training data. And we focus on the span-extraction MRC setting, where each answer $a_i = (a_{i,s}, a_{i,e})$ is a segment of text that appears in $p_i$, where $a_{i,s}, a_{i,e}$ denote the start and end positions of the ground-truth answer. The goal is to train a powerful model $\mathcal{M}$ on $\mathcal{D}{_s}$, and $\mathcal{M}$ can be able of performing well in other low-resource target languages.

\subsection{Base Model $\mathcal{M}$ }
 Our model is built on top of the powerful cross-lingual PLMs such as multilingual BERT and XLM-Roberta. Thereafter,  the input question $q_i$ and $p_i$ are concatenated with two special tokens \texttt{[SEP]} and \texttt{[CLS]} to form the input sequence $\mathbf{X}$ , as shown in the Figure \ref{modeloverview} (a). \texttt{[CLS]} is used to mark the begin of the input sentence and \texttt{[SEP]} is responsible for separating the passage and question. We then feed $\mathbf{X}$ into the encoder, and produce contextualized token representations $\mathcal{X} \in \mathbf{R}^{l\times d}$:
 \begin{equation}
     \mathcal{X} = \mathcal{H}(\mathbf{X})
 \end{equation}
where $\mathcal{H}$ is last encoder layer of cross-lingual PLMs, $l$ is the max length of input sequence and $d$ is the vector dimension of each token, separately. 

Then, to predict the start position and end position of the correct answer span in $\mathbf{X}$, the probability distributions are induced over the entire sequence by feeding $\mathcal{X}$ into a linear classification layer and followed by a softmax function.
\begin{equation}
  \mathcal{P}(s=i|\mathbf{X}), \mathcal{P}(e=i|\mathbf{X})=  \texttt{softmax}(\mathcal{W} \cdot \mathcal{X}^{T} +b) 
\label{eq:mrc}
\end{equation}
where $\mathcal{W} \in \mathbf{R}^{2\times d}$. In the typically supervised setting, we can train a model $\mathcal{M}$ by  optimizing the following function given the input $q_i$ and $p_i$:
\begin{equation}
\begin{aligned}
    \mathcal{L}_{mrc} & =
     -\texttt{log}\mathcal{P}(s=a_{i,s}|\mathbf{X}) - \texttt{log}\mathcal{P}(e=a_{i,e}|\mathbf{X})
     \\
     & =
    -\texttt{log}\mathcal{P}(s=a_{i,s}|p_i,q_i) - \texttt{log}\mathcal{P}(e=a_{i,e}|p_i,q_i)
\end{aligned}
\end{equation}

Although, this approach achieves superior performances in xMRC, it only considers the top-1 predicted result while optimizing the model with cross-entropy loss, ignoring many correct predictions exits in top-k confident predictions, and thus, making the model sub-optimized. We overcome this issue  by (1) developing a hard-learning algorithm with utilizing a pre-obtained n-best prediction set, and (2) proposing an answer-aware contrastive learning mechanism to leverage hard-negatives over training. We illustrate these two strategies in the following sections, separately.

\begin{table*}[tb]
    \centering \scriptsize
    \begin{tabular}{l}
    \toprule
        \textbf{An example from MLQA training dataset.}  \\
    \midrule
        \textbf{Question:} To whom did the Virgin Mary allegedly appear in 1858 in Lourdes France? \\
        \textbf{Passage:} Architecturally, the school has a Catholic character. Atop the Main Building's gold dome is a golden statue of the Virgin Mary. 
        Immediately in front of the Main \\ 
        Building  and facing it, is a copper statue of Christ with arms upraised with the legend "Venite Ad Me Omnes".  Next to the Main Building is the Basilica of the Sacred Heart.\\ 
        Immediately  behind the basilica is the Grotto, a Marian place of prayer and reflection. It is a replica of the grotto at Lourdes, France where the Virgin Mary reputedly appeared \\
        to \textbf{Saint Bernadette  Soubirous} in 1858.  At the end of the main drive (and in a direct line  that connects through 3 statues and the Gold Dome), is a simple, modern stone statue \\ of Mary. 
        \\
        \textbf{Answer:} \textbf{Saint Bernadette  Soubirous} \\
        \textbf{$\mathcal{Z}$ (top-k predictions) :} Saint Bernadette,    Bernadette, \textbf{Saint Bernadette  Soubirous}, Saint, the Virgin Mary reputedly appearedto Saint Bernadette  Soubirous,..., Saint...  \\

     \bottomrule
\end{tabular}
\caption{
    Examples of the input, answer text and $Z$. The correct answer is in bold. And in this example, the correct answer occurs in top-3 predictions from the model.
} 
\label{tab:formulation}
\vspace{-8pt}
\end{table*}

\subsection{Hard-Learning Algorithm}

In this component, we aim to  develop a hard-Learning (HL) algorithm during fine-tuning to maximize the likelihood for the accurate answer to be included in the set of top  k  predicted results, which is from  pre-obtained highly  confident  predictions of a basic model. That is, HL enables the model to focus on the spans which are similar with the ground-truth answer to achieve the goal of recall, as shown in Figure  \ref{modeloverview} (b).

\paragraph{Definition} Inspired by \cite{DBLP:conf/emnlp/MinCHZ19}, we define the correct answer for each question as a particular derivation that a model is required to solve for the answer prediction. Given a question $q_i$ and a passage $p_i$, 
Let $\mathcal{Z}$ = $\{ z_1, z_2,...,z_{k-1}, a_i\}$ be the set which contains top-k possible predictions from a baseline model (i.e., XML-R). Seen from Table \ref{tab:formulation}, we assume it contains an unique correct answer\footnote{If the correct answer doesn't occur in top-k predictions, we will replace the last one in $\mathcal{Z}$ with the correct answer. } ($a_i$)  that the model
 wants to learn to find, and potentially  other ones which are hard to classify.

\paragraph{Algorithm} 
In our work, the model not only has access to $q_i$ and $p_i$, but also $\mathcal{Z}$, and then we assume that  each $z_l$ in $\mathcal{Z}$ can be seen as the  \textit{true} solution for the given question. Intuitively, we compute the maximum marginal likelihood (MML) to marginalize the likelihood of each $z_l \in \mathcal{Z}$ given the $q_i$ and $p_i$, and the model can be optimized by the following loss function:
\begin{equation}
    \mathcal{L}_{mml} = -\texttt{log}\sum_{z_l \in \mathcal{Z}}\mathcal{P}(z_l|q_i,p_i)
\end{equation}

However, directly using MML for optimization can cause the model to fit on the noisy span labels contained in $\mathcal{Z}$. Specifically, in our settings, elements in $\mathcal{Z}$ can be categorized into three types: (1) ground-truth answer span (only 1); (2) spans which only match the start position or end position of the correct one; (3) spans which mismatch both the start and end positions. For the latter two types, the model is supposed to  give lower or even zero probability.
For example, as presented in Table \ref{tab:formulation}, top-1 prediction ``Saint Bernadette'' and top 4 prediction ``Saint'' match the start position of the correct one but mismatch the end position.
But when minimizing MML, it can assign high probabilities to any element in $\mathcal{Z}$.
To address this problem, we utilize a HL algorithm where different weights are assigned to each element from $\mathcal{Z}$.
Then the model can be optimized via a re-written standard cross-entropy loss as:
\begin{equation}
\begin{aligned}
     \mathcal{L}_{Hard} &=  
     -\sum_{l=1}^{k} w_l\texttt{log}\sum_{z_l \in \mathcal{Z}}\mathcal{P}(z_l|q_i,p_i)
     \end{aligned}
\end{equation}
where  $w_l$ is the learnable parameter.

\subsection{Answer-Aware Contrastive Learning}
HL encourages the model to focus on the spans which are similar with the ground-truth answer. Then, we deploy an answer-aware contrastive learning (AA-CL) mechanism to target on precision at top-1, which constructs hard-negatives using the most likely prediction with the ground-truth answer at each update to learn noise-invariant representations, which is complementary with the proposed HL algorithm. Therefore, to correctly identify hard-negatives, its relationship with the positive (ground-truth answer) must be carefully reasoned by the model, as shown in Figure \ref{modeloverview} (c).
In particular, we get the top-k answer predictions $\mathcal{A} = \{a_i^{1}, a_i^{2},..., a_i^{k}\}$ in each back propagation. And we select the one of $\mathcal{A}$ as the hard-negative example, which has the largest similarity with the ground-truth answer $a_i$\footnote{We present more detailed analysis in the Appendix.}. This can be seen as an effective coarse-to-fine negative selection strategy. Formally:
 \begin{equation}
    \mathrm{a}_i^{(l)} =\mathcal{F}(\mathcal{H}(a_i^{(l)}))
 \end{equation}
 \begin{equation}
     \hat{a} = \texttt{max}_{a_i^{l} \in \mathcal{A}}\Psi(\mathrm{a}_i^{l},\mathrm{a}_i) 
 \end{equation}

where $\Psi$(,) denotes the cosine similarity function and $\mathcal{F}$ means the mean-pooling operation.  In this work, we argue that the similarity between the input question  and the ground-truth answer is higher than others in $\mathcal{A}$. Hence, we can get the positive question-answer pair ($q_i$,$a_i$) and hard-negative pair ($q_i$,$\hat{a}$). For each pair, we use the contrastive objective to establish their correspondence among them in a shared semantic latent space:
\begin{equation}
   \mathcal{L}_{\text{contrast}}= -\text{log} \frac{\text{exp}(\Psi (\mathbf{r}_{q},\mathbf{r}_{a(pos)})/\tau)}{\sum_{n=1}^{B} \text{exp} (\Psi (\mathbf{r}_{q},\mathbf{r}_{a(n)})/\tau)},
\label{eq:contrast}
\end{equation}
where $B$ and $\tau$ are mini-batch and temperature. $\mathbf{r}_q$ and $\mathbf{r}_{a(pos)}$ denote the representation of question $q_i$ and $a_i$.
By this means, unlike only selecting negatives randomly or  in-batch negatives, we also introduce the hard-negatives from high confidence predictions of the model over training, and thus, $\mathcal{M}$ can obtain the coarse-to-fine presentations in token-level. During the fine-tuning, $\mathcal{M}$  is optimized via $\mathcal{L}_{\text{Hard}}$ and  $\mathcal{L}_{\text{contrast}}$ with the weighted ratio:
\begin{equation}
    \mathcal{L}_{\text{final}} = \alpha \mathcal{L}_{\text{contrast}} + (1-\alpha) \mathcal{L}_{\text{hard}}
    \label{loss}
\end{equation}

\begin{table}[t]
    
    \begin{center}
    \tiny
    \begin{tabular}{l|cccccccc}
    \toprule
    \multirow{2}{*}{\textbf{Dataset}} & \multirow{2}{*}{\textbf{Train}} &\multirow{2}{*}{\textbf{Dev}} & \multicolumn{6}{c}{\textbf{Test}} \\
    \cmidrule{4-9}
    & & & en & es & de & ar & hi & vi \\ 
    \midrule
    XQUAD  & 87,599 & - & 1,190  & 1,190& 1,190 &  1,190   &  1,190  &  1,190  \\
    MLQA        & 87,599 & 1,148 & 11,590 & 5,253 & 4,517 & 5,335 & 4,918 & 5,495   \\
    \bottomrule
    \end{tabular}
    \end{center}
    \caption{The Statistics of the Datasets.} \label{tab_datasts}

\end{table}

\begin{table*}[t]

\tiny
    \begin{center}
    \begin{tabular}{ll|cccccc|c}
    \toprule
    \textbf{Setting} & \textbf{Models} & \textbf{en} & \textbf{es} & \textbf{de} & \textbf{ar} & \textbf{hi} & \textbf{vi} &  \textbf{Avg.}\\
    \midrule
    \multirow{4}{*}{zero-shot} & m-BERT & 77.70\,/\,65.30 & 64.30\,/\,46.60 & 57.90\,/\,44.30 & 45.70\,/\,29.80 & 43.80\,/\,29.70 & 57.10\,/\,38.60 &  57.80\,/\,42.40 \\
   & XLM & 74.90\,/\,62.40 & \textbf{68.00}\,/\,\textbf{49.80} & 62.20\,/\,47.60 & 54.80\,/\,36.30 & 48.80\,/\,27.30 & 61.40\,/\,41.80 &  61.70\,/\,44.20 \\
   
    & XLM-R$_{base}$  & 77.86\,/\,64.24 & 66.18\,/\,48.30 & 60.82\,/\,46.43 & 55.20\,/\,35.14 & 59.93\,/\,41.93 & 64.89\,/\,42.36 &  64.14\,/\,46.00 \\
   &Info-XLM  &79.15\,/\,64.80 & 67.07\,/\,48.49 & 58.24\,/\,46.00 & 55.15\,/\,38.78 & 59.66\,/\,43.98 & 64.44\,/\,43.28 &  64.25\,/\,47.60 \\
     & \textbf{Ours}  & \textbf{79.03}\,/\,\textbf{65.59} & 67.52\,/\,49.56 & \textbf{62.98}\,/\,\textbf{48.70} & \textbf{57.68}\,/\,\textbf{39.40} & \textbf{61.79}\,/\,\textbf{44.70} & \textbf{66.74}\,/\,\textbf{45.34} & \textbf{66.00}\,/\,\textbf{48.88} \\
    \midrule
    \multirow{4}{*}{translate-train} 
    
    & XLM-R$_{base}$  & 77.15\,/\,64.41 & 68.51\,/\,50.40 & 62.24\,/\,47.99 & 56.60\,/\,38.42 & 61.39\,/\,43.93 & 66.70\,/\,44.36 &  65.45\,/\,48.25 \\
   
     & LAKM & 80.00\,/\,66.80 & 65.90\,/\,48.00 & 60.50\,/\,45.50 & - & - & - & - \\
   &  CalibreNet  & 79.68\,/\,66.51 & 68.04\,/\,50.77 & 61.66\,/\,47.55 & 56.14\,/\,37.83 & 59.97\,/\,43.84 & 66.92\,/\,46.59 &  65.40\,/\,48.84 \\
    & \textbf{Ours}  & \textbf{80.11}\,/\,\textbf{66.84} & \textbf{69.04}\,/\,\textbf{51.20} & \textbf{64.58}\,/\,\textbf{49.75} & \textbf{58.54}\,/\,\textbf{41.03} & \textbf{62.77}\,/\,\textbf{46.54} & \textbf{67.92}\,/\,\textbf{47.19} &  \textbf{67.16}\,/\,\textbf{50.44} \\

    \bottomrule
    \end{tabular}
    \end{center}
\caption{The overall evaluation results (F1/EM) on the MLQA  dataset.}\label{tab_main_mlqa}
\end{table*}

\begin{table*}[t]

\small
    \begin{center}
    \tiny
    \begin{tabular}{ll|cccccc|c}
    \toprule
    \textbf{Setting} & \textbf{Models} & \textbf{en} & \textbf{es} & \textbf{de} & \textbf{ar} & \textbf{hi} & \textbf{vi} &  \textbf{Avg.}\\
    \midrule
    \multirow{4}{*}{zero-shot} & 
    M-BERT & 81.50\,/\,71.20 & 75.50\,/\,56.90 & 70.60\,/\,54.00 & 61.50\,/\,45.10 & 59.20\,/\,46.00 & 69.50\,/\,49.10 &  69.63\,/\,53.72 \\
    &XLM & 81.30\,/\,68.80 & 75.60\,/\,56.90 & 72.60\,/\,55.50 & 62.60\,/\,43.20 & 63.10\,/\,46.00 & 70.40\,/\,48.70 & 70.93\,/\,53.18 \\
    & XLM-R$_{base}$  & 83.66\,/\,72.48 & 77.00\,/\,60.87 & 74.40\,/\,58.40 & 63.00\,/\,47.80 & 68.70\,/\,53.70 & 74.50\,/\,54.00 &  73.54\,/\,57.55 \\
     &Info-XLM  &\textbf{85.15}\,/\,72.80 & 76.15\,/\,59.30 & 73.88\,/\,59.00 & 63.51\,/\,49.78 & 69.66\,/\,54.90 & 73.21\,/\,55.25 &  73.76\,/\,58.51 \\
     
     & \textbf{Ours} & \textbf{84.51}\,/\,\textbf{74.59} & \textbf{78.25}\,/\,\textbf{61.67} & \textbf{75.89}\,/\,\textbf{59.79} & \textbf{65.18}\,/\,\textbf{50.04} & \textbf{70.79}\,/\,\textbf{55.45} & \textbf{75.74}\,/\,\textbf{56.46} & \textbf{75.06}\,/\,\textbf{59.87} \\
    \midrule
    \multirow{4}{*}{translate-train} 
    
    & XLM-R$_{base}$  & 82.59\,/\,71.30 & 78.55\,/\,60.20 & 76.42\,/\,60.69 & 65.15\,/\,48.42 & 71.35\,/\,56.43 & 76.10\,/\,56.68 &  75.03\,/\,58.96 \\
     & mixMRC & 82.40\,/\,69.20 & 78.80\,/\,58.70 & 75.40\,/\,58.20 & 63.60\,/\,42.40 & 66.20\,/\,50.00 & 72.60\,/\,52.70 & 73.17\,/\,55,20\\
   &  LBMRC  & 83.40\,/\,70.10 & 80.00\,/\,59.60 & 76.50\,/\,59.80 & 65.00\,/\,44.50 & 67.40\,/\,52.00 & 74.60\,/\,55.50 &  74.48\,/\,56.92 \\
    &  \textbf{Ours}  & \textbf{84.06}\,/\,\textbf{73.11} & \textbf{80.04}\,/\,\textbf{61.68} & \textbf{77.88}\,/\,\textbf{62.48} & \textbf{66.54}\,/\,\textbf{50.34} & \textbf{73.77}\,/\,\textbf{58.90} & \textbf{77.64}\,/\,\textbf{57.49} &  \textbf{76.66}\,/\,\textbf{60.66} \\
    \bottomrule
    \end{tabular}
    \end{center}
\caption{The overall evaluation results (F1/EM) on the XQUAD  dataset.  }\label{tab_main_xquad}
\end{table*}

\section{Experiments}
\subsection{Datasets and Evaluation Metrics}
We evaluate our proposed methods on two popular datasets, MLQA \cite{DBLP:conf/acl/LewisORRS20} and XQUAD \cite{DBLP:journals/corr/abs-1809-03275}, to examine the effectiveness.

\textbf{MLQA} is a popular xMRC benchmark, which covers various languages.
We evaluate our methods on six  languages: 
 including \textit{English, Arabic, German, Spanish, Hindi, Vietnamese}.

\textbf{XQUAD} is another dataset for evaluating the cross-lingual model performances, which consists of 11 languages. Similar to the setting above, we test our method with the same six languages in our experiments under the zero-shot and translation train setting. 
Table \ref{tab_datasts} shows the detailed statistics of the datasets.

 We use two evaluation metrics, Exact Match (EM) and Macro-averaged F1 score (F1), which are popularly used for accuracy evaluation of MRC models. F1 measures the part of the overlapping mark between the predicted answer and the ground-truth answer.  The exact match (EM) score is 1 if the prediction is exactly the same as the ground truth, otherwise 0.

\subsection{Implementation Details}
We build our model on XLM-R$_{base}$  on top of the Hugging Face Transformers\footnote{https://github.com/huggingface/transformers}, which contains 12 transformer layers. We use AdamW \cite{loshchilov2017decoupled} as our model optimizer, and the weight decay is set to $0.01$ for both datasets. 
The learning rate is set to 3e-5. The size of $|\mathcal{Z}|$ and  $|\mathcal{A}|$ are  20 and 50 in experiments, respectively\footnote{Intuitively, with the size of $|\mathcal{A}|$ increasing, the model achieves better performance as it can mine more $hard$ negatives. Meanwhile, it also raises the computation cost. Empirically, we choose the  $|\mathcal{A}|$ as 50 in our experiments.}. During fine-tuning, we empirically set the max input sequence length to 384. The question max length is 64. We also use the warm-up proportion and set to 0.1. The $\tau$ in Eq.\ref{eq:contrast} and batch size are 10 and 32, respectively. The $\alpha$ in Eq.\ref{loss}  is set to 0.5 in our experiments. We train the  model using 8 NVIDIA V100 GPUs with 32 GB memory for each training language data with 8 epochs and  save a checkpoint every 1000 steps.


\subsection{Baselines}
We compare our model with the following strong baselines: (1) M-BERT \cite{pires2019multilingual}, a cross-lingual version of BERT  trained on 104 parallel languages and demonstrated highly competitive in  multilingual language understanding tasks at zero- and few-shot settings; (2) XLM \cite{DBLP:conf/nips/ConneauL19}, another effective pre-trained multilingual language model achieving promising results on various cross-lingual tasks; (3) LAKM, a pre-trained task proposed by \citet{DBLP:conf/acl/YuanSBGLDFJ20} via introducing extra parallel corpus for phrase level MLM; (4) mixMRC, a translation-based data augmentation strategy developed by \citet{DBLP:conf/acl/YuanSBGLDFJ20} for xMRC; (5) LBMRC, a novel augmentation approach \cite{DBLP:conf/coling/LiuSPGYJ20} based on knowledge distillation; (6) CalibreNet \cite{DBLP:conf/wsdm/LiangSPGZJ21}, a recent model aiming to enhance the boundary detection capability of PLMs in multilingual sequence labeling task; and (7) Info-XLM \cite{DBLP:conf/naacl/ChiDWYSWSMHZ21}, a new state-of-the-art information-theoretic cross-lingual pre-training model. For fair comparisons, we use XLM-R$_{base}$ as our backbone architecture in this work.

\begin{figure*}

    \centering
    \includegraphics[width=0.9\linewidth]{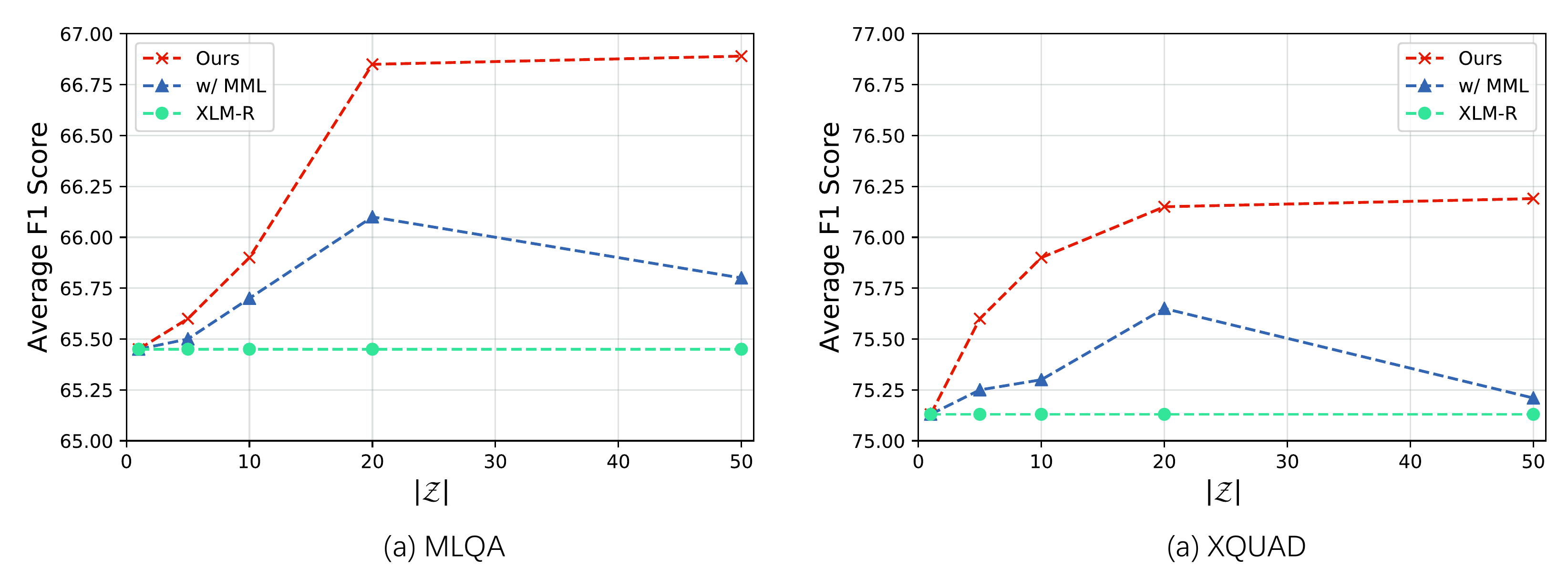}
  
    \caption{Model performances with different size of $\mathcal{Z}$ in training at the \textit{translate-train} setting on both two datasets. We use the average F1 score of six languages as the evaluation metric.}
    \label{fig:z}
 
\end{figure*}

\subsection{Results}
We compare our methods with the strong baselines in the two settings. The first is \textit{zero-shot}: we fine-tune the state-of-the-art models in English only, and then test on the English and other five low-resource languages. The second is \textit{translate-train}: we train the models by combining the translated data of all languages jointly during fine-tuning.

\paragraph{Results on MLQA} In the first set of our experiments, we evaluate various baselines on the MLQA dataset, and the results are listed in Table \ref{tab_main_mlqa}. We make several observations from the results. First, our method outperforms all baselines in all languages at the \textit{zero-shot} setting, indicating the effectiveness of our model. For instance, ours improves XLM-R$_{base}$ from $64.14\%$ to $66.00\%$ in F1 and from $46.00\%$ to $48.88\%$ in EM score on average.  Moreover, in the \textit{translate-train} setting, our approach achieves the best results $67.16\%$ and $50.44\%$ in F1/EM scores, respectively, which surpasses the strong baselines by a large margin. Third, compared with the LAKM and CalibreNet, which both utilize extra cross-lingual corpora, our model also obtains better results. Last, our model in the \textit{zero-shot } setting even outperforms XLM-R$_{base}$ in the \textit{translate-train} setting. This confirms the effectiveness of the proposed Hard-Learning algorithm and Answer-Aware Contrastive Learning.

\paragraph{Results on XQUAD} 
In order to show the generality, we  also evaluate our approach on other common used xMRC benchmark called XQUAD in six languages. 
The experimental results  are reported in Table \ref{tab_main_xquad}, which are also under the \textit{zero-shot} and \textit{translate-train} settings. Clearly, our method consistently outperforms the strong baselines in both settings. Specifically, our best model outperforms XLM-R$_{base}$ in the \textit{translate-train} setting with a clear margin in both F1 and EM scores.
In the \textit{zero-shot} setting, our model also obtains on average $1.52\%$ and $2.32\%$ improvement F1 and EM scores, respectively, in those languages. Even compared with other strong baselines like mixMRC and LBMRC, ours also show its superiority. 
The evaluation results on XQUAD further verify the effectiveness and robustness of our method. 

\section{Analysis}

In this section, we  conduct a series of ablation studies and analysis to better understand what contributes to the performance advantages of our model. Furthermore, we present the ablation study of hyper-parameter $\tau$ and AA-CL in Appendix A and B.

\subsection{Key Components}

\begin{table}[]
\footnotesize
    \centering

    \begin{tabular}{l|cccc}
    \toprule
       Models   & es & ar & vi  \\
       \midrule
      Ours   & 69.04/51.20&58.54/41.03& 67.92/47.19\\
      \midrule
      - \textit{HL} &67.64/49.45&57.10/39.18&66.00/45.49\\
      - \textit{AA-CL} &67.70/50.12&57.66/39.80&67.00/46.07\\
      w/ \textit{MML} &68.47/50.21&57.46/40.00&66.89/46.01\\
     \bottomrule
    \end{tabular}
    \caption{Ablation study of our methods on MLQA dataset at \textit{translate-training} setting. We evaluate each method in three languages: \textit{Spanish, Arabic and Vietnamese.}}

    \label{tab:component}
\end{table}

\begin{figure*}

    \centering
    \includegraphics[width=0.95\linewidth]{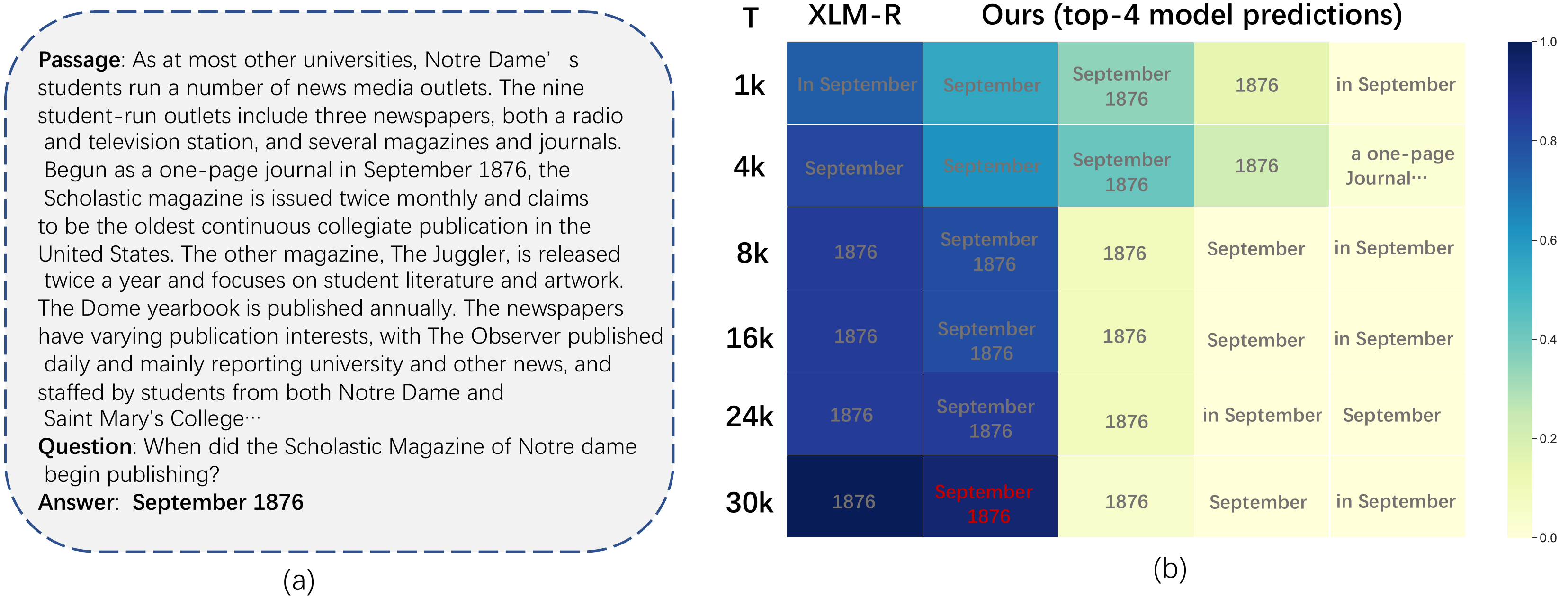}
  
    \caption{An example from MLQA dataset, with its ground-truth answer ``September 1876". For each iteration step T, we present top-1 prediction from the baseline (XLM-R$_{base}$) and top 4 predictions from ours at the \textit{translate-train} setting.}
    \label{fig:prediction}
  
\end{figure*}

  
  

To evaluate the effectiveness of our model, we conduct ablation studies by removing each key component individually. As shown in Table \ref{tab:component}, there is a obvious performance gap when removing \textit{HL}, indicating that pre-obtaining a set of predictions and training a model through hard updates play an important part in performance. Then, removing \textit{AA-CL}, the model performance drops inevitably. The results demonstrate the effectiveness of this coarse-to-fine method for utilizing hard-negatives from high confident predictions over training. In general, each key component contributes to the performance improvement of the model. In Table \ref{tab:component} we provide the results using MML as our training objective. The model performance drops about 1$\%$ in F1 and EM scores on three languages, indicating the effectiveness of HL algorithm once again.

\subsection{Size of $\mathcal{Z}$}

To assess how the proposed hard learning algorithm works with respect to the size  of pre-obtained predictions set ($|\mathcal{Z}|$), we conduct a series of experiments on both datasets with $|\mathcal{Z}| = \{ 1,5,10,20,50\}$. Figure \ref{fig:z} shows the results. For fair comparisons, AA-CL is removed in this experiment. Figure \ref{fig:z} shows that our proposed method outperforms MML and the baseline consistently with different values of $|\mathcal{Z}|$. When $|\mathcal{Z}|$ is set to 20 and 50, the model achieves comparable performances on the two datasets. Considering the computation efficiency and memory cost, we choose $|\mathcal{Z}|= 20$ in our main experiments. 
\begin{figure}
    \centering
    \includegraphics[width=0.95\linewidth]{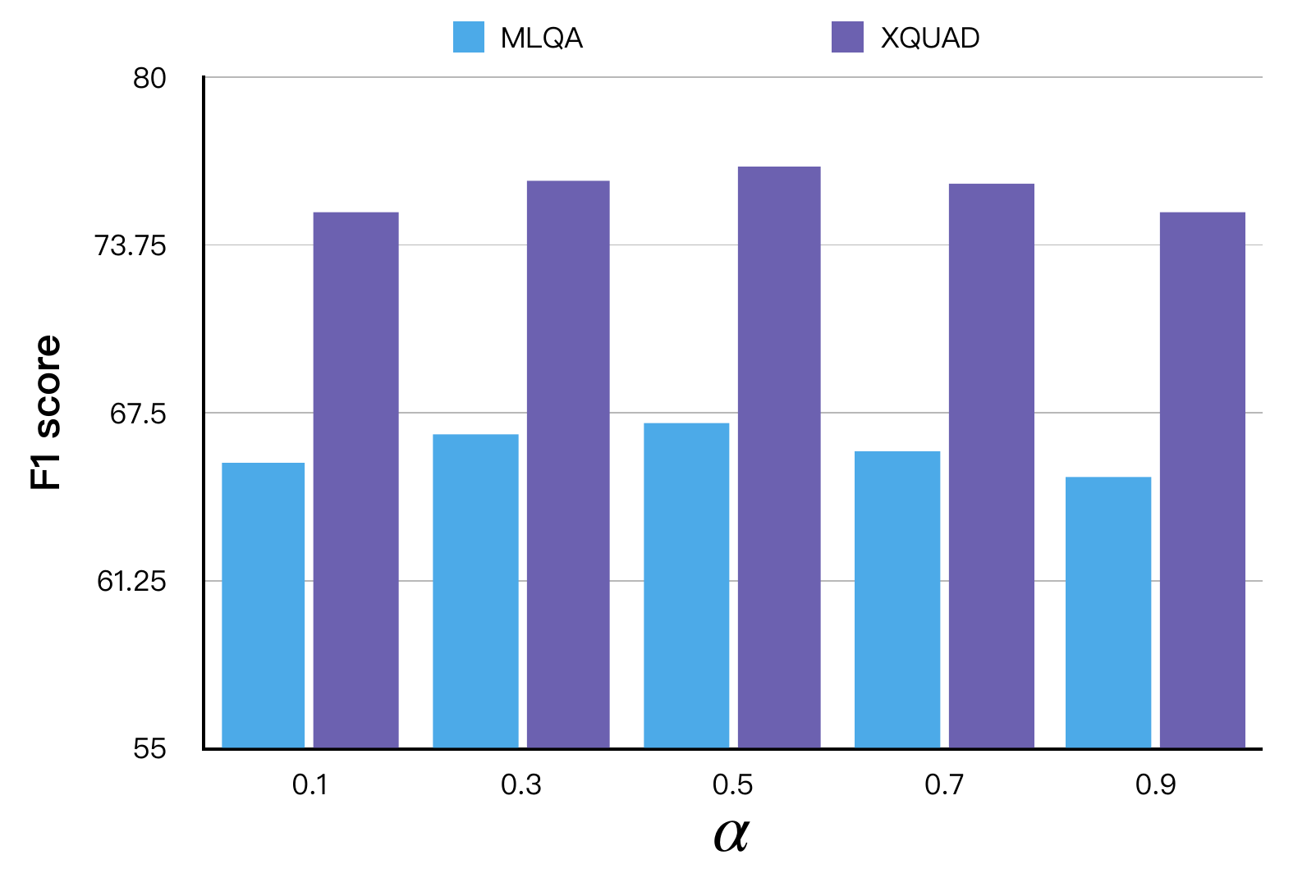}
  
    \caption{Model performances with different $\alpha$ in training at the \textit{translate-train} setting. We use the average F1 score of six languages as the evaluation metric.}
  
    \label{fig:a}
\end{figure}

\subsection{Case Study of Model Predictions Over Training}
To show how our model performs during the training process, we analyze the top predictions and assigned likelihood from the models with respect to different iteration steps (from 1k to 30k). Figure \ref{fig:prediction}  shows that both the baseline and our model first begin by assigning higher probabilities to wrong predictions, like ``in September" and "September", but gradually our method leans to favor the true prediction. Unfortunately, XLM-R$_{base}$ still insists on making the wrong prediction until the end of the training, indicating that it may confused by the similar spans with the correct answer (``1876" vs. ``September 1876"), which can be seen as an understandable mistake. The visualization in Figure \ref{fig:prediction} (b) shows the ability of our model in identifying the correct answer from many similar spans.


\subsection{Hyper-parameter $\alpha$}
It is essential to study the sensitivity analysis of  $\alpha$, since we train our model in a multi-task manner.
Thereafter, we conduct additional experiments to study the effect of different values of $\alpha$ on optimizing the model on both datasets. We test the model performance with  $\alpha \in \{0.1,0.3,0.5,0.7,0.9\}$. From Figure \ref{fig:a}, we find that the model performances on MLQA and XQUAD show similar trends  $\alpha$, and our method achieves the best results while $\alpha=0.5$.




\section{Conclusion}

In this paper, we tackle the challenge of 
  exploring the potential of mining useful task-related knowledge from n-best answer predictions. Concretely, we
  decompose the training for xMRC model into two stages: (1) At the first stage, 
 we target at recall at top-k predicted results, and thus, develop a hard learning algorithm to progressively encourage the model to give higher attention to the pre-obtained top-k predictions with taking these as weak supervision.
(2) Then, we propose an  answer-aware contrastive learning to  strengthen the model's ability to further distinguish the correct span from top-k possible spans to achieve the goal of  precision at top-1.
Experimental results show that our model achieves competitive performances compared to the state-of-the-art on two public benchmark datasets. The systematic analysis further demonstrates the effectiveness of each component in our model. Future work can include an extension of how to employ AA-CL to other natural language understanding tasks.

\section{Acknowledgments}
Jian Pei’s research is supported in part by the
NSERC Discovery Grant program. All opinions,
findings, conclusions and recommendations in this
paper are those of the authors and do not necessarily reflect the views of the funding agencies.


\bibliography{main.bib}

\end{document}


\maketitle

\appendix

\begin{figure}
    \centering
    \includegraphics[width=0.95\linewidth]{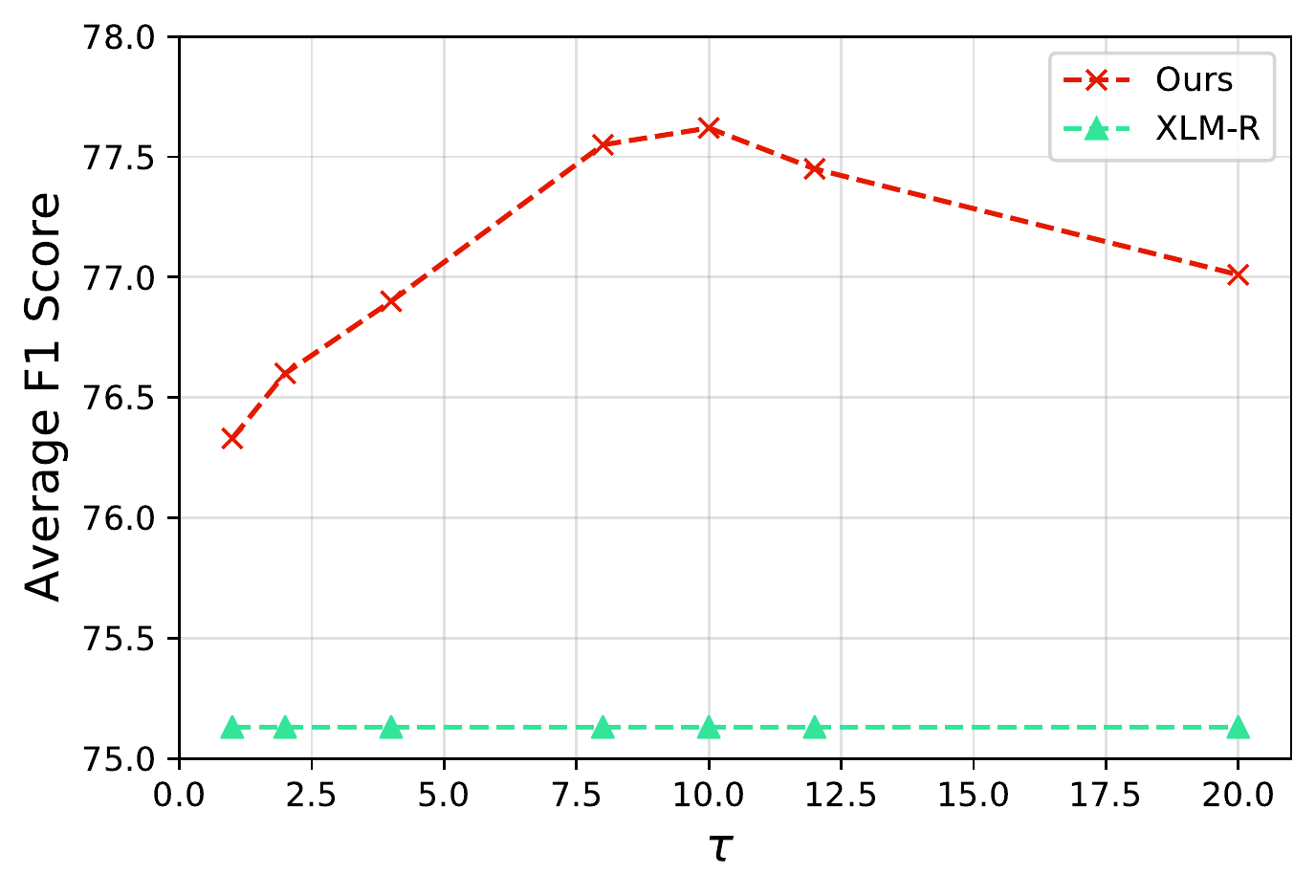}
  
    \caption{Model performances with different $\tau$ in training at the \textit{translate-train} setting. We use the average F1 score of six languages as the evaluation metric.}
    \vspace{-10pt}
    \label{fig:t}
\end{figure}

\section{A}

\subsection{Hyper-parameter $\tau$}
The choice $\tau$ of the proposed answer-aware negative contrastive learning (A-ANCL) in Eq.8 may influence the model performance, and thus, we conduct a series of sensitivity analysis experiments on the MLQA dataset. In detail, we test the model performance with using $\tau \in \{1,2,4,8,10,12,20\}$. Seen from Figure \ref{fig:t}, we can draw the following observation: Different  $\tau $ indeed contributes performance improvement or degradation. And our model can achieve best result when $\tau$ set to 10 in this work.

\section{B}

\subsection{Negatives in AA-CL}
In this component, we raise an interesting question: Why we construct hard-negatives just  using the span which is most similar with the ground-truth answer? Therefore, we conduct corresponding experiments with following settings: (1) we utilize the top $\Theta$ similar ones with the correct answer to construct hard-negatives in AA-CL, where $\Theta \in \{1,10,20\}$, (2) we choose the top 1 prediction but not equal to the correct answer as the hard-negative, (3) we randomly choose one from $\mathcal{A}$ as the hard-negative. Correspondingly, we present the detailed results in Table \ref{tab:aacl}. 
When looked into the table, we easily find that our model achieves comparable performances under the different $\Theta$ settings, showing only selecting the most similar one with the correct answer is abundant for training. Hence, considering the computation cost, we  choose   $\Theta$ = 1 in our work.
Then we also can observe that using the top 1 prediction or randomly sampling one in  $\mathcal{A}$ as hard-negative brings performance degradation in terms of F1/EM score, which proves the superiority of our hard-negative selecting strategy. 

\begin{table}[]

    \centering

    \begin{tabular}{l|cccc}
    \toprule
& \multicolumn{2}{c}{MLQA} &\multicolumn{2}{c}{XQUAD}  \\
     \cmidrule{2-5}
        Settings  & F1 & EM &F1& EM\\
       \midrule
      $\Theta$ = 1 &66.00&48.88&75.06&59.87\\
      $\Theta$ = 10 &65.80&48.67&74.96&59.50\\
      $\Theta$ = 20 &66.10&49.12&74.89&59.41\\
     \midrule
     top 1 prediction&65.02&47.94&74.55&59.24 \\
     \midrule
     random sample &65.11&48.02&74.21&58.86\\
     \bottomrule
    \end{tabular}
    \caption{Analysis on the strategy of selecting hard-negatives in AA-CL. We conduct
    our methods on MLQA and XQUAD datasets at \textit{zero-shot training} setting. And we use the average F1/EM score as evaluation metrics.}

    \label{tab:aacl}
\end{table}

\section{C}
\subsection{Comprehensive Results of analysis}

\begin{table*}[ht]
\vspace{-1em}
\tiny
    \begin{center}
    \begin{tabular}{ll|cccccc|c}
    \toprule
    \textbf{Dataset} & \textbf{Models} & \textbf{en} & \textbf{es} & \textbf{de} & \textbf{ar} & \textbf{hi} & \textbf{vi} &  \textbf{Avg.} \\
    \midrule
    \multirow{9}{*}{MLQA}
     &Info-XLM  &79.15\,/\,64.80 & 67.07\,/\,48.49 & 58.24\,/\,46.00 & 55.15\,/\,38.78 & 59.66\,/\,43.98 & 64.44\,/\,43.28 &  64.25\,/\,47.60 \\

    &XLM-A  &78.80\,/\,64.70 & 67.64\,/\,48.95 & 61.88\,/\,47.23 & 55.95\,/\,37.42 & 58.04\,/\,41.47 & 65.59\,/\,43.49 &  64.72\,/\,47.12 \\
   
   & XLM-R$_{base}$  & 77.86\,/\,64.24 & 66.18\,/\,48.30 & 60.82\,/\,46.43 & 55.20\,/\,35.14 & 59.93\,/\,41.93 & 64.89\,/\,42.36 &  64.14\,/\,46.00 \\
    & \textbf{Ours}  & \textbf{79.00}\,/\,\textbf{65.50} ($\pm$ 0.1) & \textbf{67.52}\,/\,\textbf{49.50} ($\pm$ 0.1) & \textbf{62.98}\,/\,\textbf{48.60} ($\pm$ 0.15) & \textbf{57.68}\,/\,\textbf{39.40} ($\pm$ 0.1) & \textbf{61.79}\,/\,\textbf{44.70} ($\pm$ 0.3) & \textbf{66.74}\,/\,\textbf{45.44} ($\pm$ 0.2) & \textbf{66.00}\,/\,\textbf{48.88} ($\pm$ 0.2) \\

    & Removing \textit{HL} &78.10\,/\,64.50 & 66.44\,/\,48.00 & 62.40\,/\,47.20 & 55.90\,/\,38.68 & 60.00\,/\,43.90 & 65.36\,/\,44.00 &  64.71\,/\,47.50 \\
    &Removing \textit{AA-CL} & 78.10\,/\,64.00 &66.40\,/\,49.32 & 62.29\,/\,48.25 & 56.36\,/\,38.00 & 60.30\,/\,43.93 & 65.70\,/\,44.27 & 64.86\,/\,47.63 \\
    &w/ \textit{MML}  & 78.46\,/\,64.00 & 67.27\,/\,48.71 & 61.90\,/\,46.90 & 56.26\,/\,38.50 & 60.50\,/\,44.21 & 65.69\,/\,44.51 &  65.06\,/\,47.80 \\
    \midrule
    \multirow{6}{*}{XQUAD} 
     &Info-XLM  &\textbf{85.15}\,/\,72.80 & 76.15\,/\,59.30 & 73.88\,/\,59.00 & 63.51\,/\,49.78 & 69.66\,/\,54.90 & 73.21\,/\,55.25 &  73.76\,/\,58.51 \\
      & XLM-A  &83.50\,/\,72.30 & 77.57\,/\,60.45 & 74.45\,/\,59.30 & \textbf{66.02}\,/\,\textbf{51.18} & 69.80\,/\,54.80 & 73.20\,/\,55.80 &  74.04\,/\,58.61 \\
     & \textbf{Ours} & \textbf{84.50}\,/\,\textbf{74.50} ($\pm$ 0.2) & \textbf{78.25}\,/\,\textbf{61.60} ($\pm$ 0.3) & \textbf{75.90}\,/\,\textbf{59.80} ($\pm$ 0.1) & 65.18\,/\,50.04 ($\pm$ 0.2) & \textbf{70.79}\,/\,\textbf{55.40} ($\pm$ 0.25) & \textbf{75.70}\,/\,\textbf{56.40} ($\pm$ 0.15) & \textbf{75.06}\,/\,\textbf{59.87} ($\pm$ 0.2)\\
     & Removing \textit{HL} &83.23\,/\,71.70 & 76.14\,/\,60.45 & 74.60\,/\,58.70 & 64.10\,/\,49.18 & 69.70\,/\,54.00 & 73.87\,/\,55.49 &  73.61\,/\,58.25 \\
    &Removing \textit{AA-CL} & 83.40\,/\,71.89 &76.21\,/\,60.01 & 74.79\,/\,58.88 & 64.66\,/\,49.38 & 70.01\,/\,54.01 & 74.00\,/\,55.67 & 73.84\,/\,58.31 \\
    &w/ \textit{ML}  & 83.05\,/\,73.08 & 77.17\,/\,60.61 & 74.40\,/\,58.59 & 64.46\,/\,48.40 & 69.77\,/\,53.99 & 73.89\,/\,55.41 &  73.79\,/\,58.40 \\
    \bottomrule
    \end{tabular}
    \end{center}
\caption{Ablation study of our methods  at \textit{zero-shot} setting. The overall evaluation results (F1/EM) on both two datasets. We re-train the new baselines in our local environment. }\label{tab_main_xquad}

\end{table*}